# An Ergonomic Interaction Workspace Analysis Method for the Optimal Design of a Surgical Master Manipulator

D. Zhang, J. Liu, G.-Z. Yang

*The Hamlyn Centre for Robotic Surgery, Imperial College London*

## INTRODUCTION

The master manipulator of a master control console, is used to interact and capture the gestures from an operator, and control the slave robot remotely [1]. Several fundamental aspects for designing a master manipulator should be taken into account, such as safety, maneuverability and accuracy[2]. Master control console is a place where robots collaborate with humans in a shared environment. Therefore, ergonomics is also an important aspect. With ergonomic consideration, the operator can feel more comfortable and confident to conduct the surgical tasks with higher efficiency, and the quality of the tele-operational robotic surgery can be improved.

Human reference model of the Master Motor Map (MMM) framework has been proposed for human arm workspace modelling and analysis [3]. However, for robotic surgery application, the effective workspace can be more specific based on the characteristics of master-slave operational mode. Ergonomic body posture of the surgeon during robot-assisted surgery was analyzed for the da Vinci console [4]. But its main focus was on geometric data and ergonomic instructions for an optimal console setting without considering the master manipulators. In this paper, an Ergonomic Interaction Workspace Analysis method is proposed to optimize master manipulators and fulfil ergonomics consideration for remote robotic surgery.

## MATERIALS AND METHODS

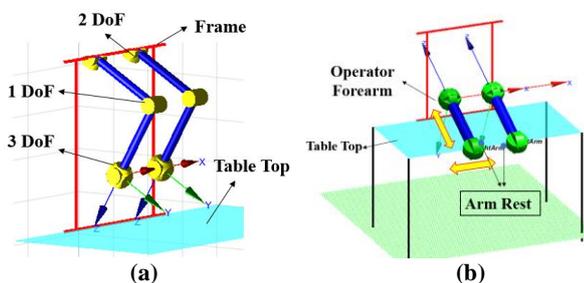

**Figure 1.** (a) Master Manipulators; (b) Equivalent Human Ergonomic Model (EHEM).

As shown in Fig.1-(a), a pair of six Degrees of Freedom (DoF) master manipulators are taken for the calculation in this paper as an example. To control the master manipulators, operators normally rest their arm on the arm rest and hold the end-effectors of the manipulators via fingers. General motions include moving along the edge of the desk, moving arms forward and backward, pitch and roll motions on the fulcrum of the arm rest, and three dimensional wrist motions. Based on the general movements mentioned above, a pre-defined Equivalent Human Ergonomic Model (EHEM) can be built up for analysis of human comfortable workspace (Fig.1-(b)), with ergonomics consideration.

As shown in Fig.2-(a), reachable workspace of the master manipulators and the EHEM are analyzed at first, while Master Manipulator Dexterous Workspace (MMDW) is calculated based on the manipulability measurement index (see Fig.2-(b)), which was introduced by Yoshikawa [5]. It describes the distance of a given pose of the manipulator to a singular configuration. Joint limits have significant impacts on the end effector's maneuverability in workspace, and the influence can be addressed by adding a penalization term to augment the Jacobian matrix of the master manipulator [6]. The maneuverability can be written as follows:

$$Dex = \sqrt{\det(JJ^T)}[1 - \exp(-K \prod_{j=1}^{N} \frac{(\theta_j - \theta_{j,down})(\theta_{j,up} - \theta_j)}{(\theta_{j,up} - \theta_{j,down})^2})] \quad (1)$$

where $J$ is the Jacobian matrix, $K$ is a scaling factor, $N$ is the number of joints, $\theta_j$ is the joint angle, $\theta_{j,up}$ and $\theta_{j,down}$ are the up and down boundary of joint limitation respectively.

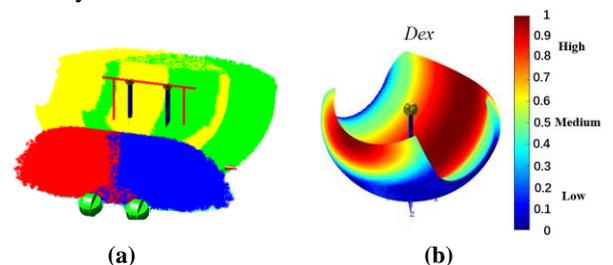

**Figure 2.** (a) Master Robot Reachable Workspace; (b) Master Robot Dexterous Workspace.

It is difficult to build a precise model for dexterity calculation, so Monte Carlo based technique is used for estimation. A large set of random master robot arm configurations are generated, and the position and orientation of the end-effector are computed via forward kinematics with corresponding dexterity value $Dex$ after normalization of the calculation results from equation (1). The scatter data of the calculation results are written as $Dex = S(x, y, z) \in \square_s$, $(p_{Lx} \le x \le p_{Ux}, p_{Ly} \le y \le p_{Ly}, p_{Lz} \le z \le p_{Uz})$. Some points of low dexterity value are abandoned based on a threshold value. Fig.3-(a) demonstrates the isosurface with isovalue of 0.3, which is a segmentation surface of low and high dexterity. As for workspace representation, the workspace can be divided into equally sized cubes called voxels, with a predefined desired resolution [7]. The voxel value can be represented as reachability by filling with dexterity measurement information describing the capabilities of the slave robot. The corresponding dexterity value $Dex$ is mapped a to discretized workspace with cubic voxel, and can be represented as $Dex' = V(i,j,k) \in \square_v$, $(1 \leq i \leq \frac{(p_{Ux}-p_{Lx})}{r}, 1 \leq j \leq \frac{(p_{Uy}-p_{Ly})}{r}, 1 \leq k \leq \frac{(p_{Uz}-p_{Lz})}{r})$, where $r$ is the minimum distance between two scattered points. The corresponding relationship of the volume data and the scatter data is described as follows:

$$V(i,j,k) = S(\frac{x-p_{x\min}}{r}, \frac{y-p_{y\min}}{r}, \frac{z-p_{z\min}}{r}) \quad (2)$$

After calculating the MMDW, the shared volume between the right and the left arm and the overall dexterous workspace volume data can be generated, as shown is Fig.3-(b) based on the following equation.

$$V_{dual}(i',j',k') = \alpha[V(i-r_x, j-r_y, k-r_z) + V(i+D-r_x, j-r_y, k-r_z)] \quad (3)$$

where $R = (r_x, r_y, r_z)$ is the transition vector of the frame of the operator and the master robot, $D$ is the distance between the two master manipulator along the fixed frame and $\alpha$ is the standardization factor.

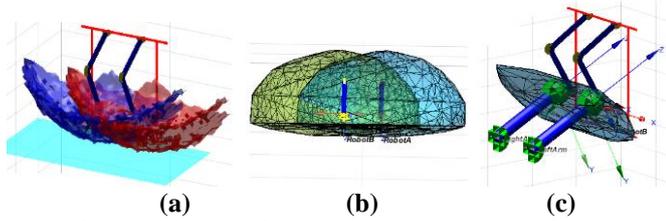

(a)      (b)      (c)

**Figure 3. (a) Isosurface of Dual Arms Dexterous Workspace with isovalue of 0.3; (b) Visualization of Overlap Volume of Dual Arm Dexterous Workspace; (c) Ergonomic Interaction Workspace.**

The Equivalent Human Ergonomic Index in the 3-D workspace is represented as $E(i',j',k') = 0.5 \in \square_v$. The master manipulator's workspace should have the maximum intersection of the human arm's comfortable moving space. Therefore, the Ergonomic Interaction Workspace can be evaluated based on the following equation:

$$\sum_{i'=1}^{i'\max} \sum_{j'=1}^{j'\max} \sum_{k'=1}^{k'\max} [(E_{Left}(i',j',k') + E_{Right}(i',j',k'))] V_{dual}(i',j',k') \quad (4)$$

## RESULTS

**Table 1 Optimized Master Manipulator DH Parameters**

|   | 1 | 2 | 3 | 4 | 5 | 6 |
|---|---|---|---|---|---|---|
| $\alpha$ | 0 | pi/2 | -pi/2 | 0 | -pi/2 | pi/2 |
| $a$ | 0 | 0 | (0.15m)0.26m | (0.15m)0.18m | 0 | 0 |
| $d$ | 0 | 0 | 0 | 0 | 0 | 0 |
| $\theta$ | $\theta_1$ | $\theta_2$ | $\theta_3$ | $\theta_4$ | $\theta_5$ | $\theta_6$ |

The optimized solution can be obtained by changing parameters of the manipulators' link length and the distance of the base of the two master manipulators along the mounting frame. Fig.3-(c) is the visualization result of the Ergonomic Interaction Workspace. Optimization method is used to maximum the volume of the Ergonomic Interaction Workspace, and the final D-H parameters of the master manipulator can be obtained (Table 1), while $D$=0.20 m. The optimization index can be represented as the percentage of the volume of Ergonomic Interaction Workspace $V_I$ to the volume of Human Ergonomic Workspace $V_R$ as $F = V_I/V_R$. The original result is $F = 0.0057$, while the optimized result is $F = 0.2732$

## DISCUSSIONS

An Equivalent Human Ergonomic Model (EHEM) is defined in this paper, which represents a simplified model for surgeon to realize remote control of surgical robots with armrests in the confined environment. The dexterity of the master manipulator is calculated, which is known as the Master Manipulator Dexterous Workspace (MMDW). The Ergonomic Interaction Workspace is formed by the intersection between the MMDW and EHEM. Optimization method is used to determine the suitable link length of the master manipulators for human-robot interaction tasks, which plays an important role in maximizing the interaction dexterous workspace volume in a predefined environment to improve the outcomes of master-slave remote control during robotic surgery. After optimization, the percentage of volume of Ergonomic Interaction Workspace is improved by 50 times. Future work will includes building up a prototype and conducting user study to test the ergonomics and discover how much operational efficiency can be improved by the optimized design using Ergonomic Interaction Workspace Analysis method [8][9].